\documentclass[letterpaper, 10 pt, conference]{ieeeconf}  

\IEEEoverridecommandlockouts                              

\usepackage{graphics} 
\usepackage{epsfig} 
\usepackage{times} 
\usepackage{amsmath} 
\usepackage{amssymb}  
\usepackage[first-style=long]{acro}
\usepackage{xcolor}
\usepackage{cite}
\usepackage[10pt]{moresize}
\usepackage{siunitx}
\usepackage{mathtools,array}
\usepackage{framed}
\usepackage{tikz}

\DeclareAcronym{ocp}{
	short = OCP,
	long = optimal control problem,
}

\DeclareAcronym{rrt}{
	short = RRT,
	long = rapidly-exploring random tree,
}

\DeclareAcronym{3D}{
	short=3D,
	long=three-dimensional
}

\title{\LARGE \bf
Hierarchical Time-Optimal Planning for Multi-Vehicle Racing*
}

\author{Georg Jank, Matthias Rowold, and Boris Lohmann
\thanks{*This work was not supported by any organization}
\thanks{All authors are with the Chair of Automatic Control, Department of Mechanical Engineering, TUM School of Engineering and Design, Technical University of Munich, 85748 Garching, Germany
        {\tt\small georg.jank@tum.de}}}%

\begin{document}

\maketitle
\thispagestyle{empty}
\pagestyle{empty}

\begin{tikzpicture}[remember picture,overlay]
    \node[xshift=0mm,yshift=0.5mm ,anchor=south] at (current page.south){%
    \resizebox{\textwidth}{!}{\begin{minipage}{0.9\paperwidth}
	\centering
	\begin{framed}
		\copyright~2023 IEEE. Personal use of this material is permitted. Permission from IEEE must be obtained for all other uses, in any current or future media, including reprinting/republishing this material for advertising or promotional purposes, creating new collective works, for resale or redistribution to servers or lists, or reuse of any copyrighted component of this work in other works.
	\end{framed}
\end{minipage}}};
\end{tikzpicture}

\begin{abstract}

This paper presents a hierarchical planning algorithm for racing with multiple opponents. The two-stage approach consists of a high-level behavioral planning step and a low-level optimization step. By combining discrete and continuous planning methods, our algorithm encourages global time optimality without being limited by coarse discretization. In the behavioral planning step, the fastest behavior is determined with a low-resolution spatio-temporal visibility graph. 
Based on the selected behavior, we calculate maneuver envelopes that are subsequently applied as constraints in a time-optimal control problem. The performance of our method is comparable to a parallel approach that selects the fastest trajectory from multiple optimizations with different behavior classes. However, our algorithm can be executed on a single core. This significantly reduces computational requirements, especially when multiple opponents are involved. Therefore, the proposed method is an efficient and practical solution for real-time multi-vehicle racing scenarios. 

\end{abstract}

\section{Introduction}

Planning trajectories in environments with dynamic obstacles is a major task in autonomous driving. Although approaches for traffic scenarios and racing can be similar, high speeds, small distances, and different rules pose a unique challenge in competitive driving on race tracks (like the Indy Autonomous Challenge). Trajectory planning in this environment requires rapid solving of non-convex optimization problems to generate time-optimal behavior (e.g. left or right overtake) with a corresponding feasible trajectory.

The majority of recent planning approaches for racing solve the behavior and trajectory generation problem in one step by selecting the cost-minimum option from a finite number of generated trajectories \cite{Raji2022, Rowold2022, Ogretmen2022}. These methods are not prone to local optima, as they cover a large region of the search space. However, they only find discrete-optimal solutions, as they do not explore all possible trajectories. We call them discrete methods in the following. Numerical optimization-based methods, on the other hand, solve an \ac{ocp} with only the time or progress along a curve being discretized. Thus, they are often referred to as continuous methods. As the control problem is non-convex, they converge to different local optima, depending on the initialization. One way to consider the non-convexity is to solve multiple \acp{ocp} in parallel, one for each behavior class. However, this does not scale well for multiple opponents and relies on parallel processing capabilities to achieve low computation times.

For rapid planning in an environment with multiple opponents, we propose a hierarchical planning approach that uses a spatio-temporal visibility graph to determine a high-level behavior and set the constraints for a low-level numerical optimization. In essence, we adopt discrete methods for exploration and continuous methods for exploitation, thereby combining the strengths of both approaches.

\section{Related Work}\label{sec:related_work}
Discrete planning methods generate and compare a finite number of candidate trajectories. There are two main subcategories of discrete planning approaches: sampling-based and graph search methods \cite{LaValle2006}.
Sampling-based methods, using a \ac{rrt} \cite{Jeon2013}, generate trajectory candidates randomly with forward dynamics. These are checked for feasibility and ranked to find the discrete-optimal trajectory. Other sampling-based approaches, applied in racing, sample jerk-minimal splines \cite{Raji2022,Ogretmen2022}. A major disadvantage of sampling-based methods is the large number of candidates required to plan complex driving maneuvers~\cite{Ziegler2009}.

Graph search methods aim to reduce the number of trajectory candidates by creating a graph of feasible trajectory segments called edges. A graph search then determines the cost-minimal sequence of edges. In path-velocity decomposition \cite{Kant1986}, trivial overtaking maneuvers are planned by calculating a collision-free velocity profile on an optimal path derived from a spatial graph. Even though such approaches have been applied in racing \cite{Stahl2019}, they are not time-optimal, as they do not fully capture the spatio-temporal character of the problem. Directly considering dynamic obstacles in the graph leads to spatio-temporal graphs \cite{Rowold2022}. However, this requires at least one additional dimension (time, velocity, or both). Therefore, the discretization must be kept coarse to mitigate the curse of dimensionality.

Continuous methods only discretize the time or progress along a curve and solve an \ac{ocp} numerically. As they converge to continuous local optima, they have become a common choice for trajectory planning in autonomous motor-sport  \cite{Buyval2017,Kalaria2021,Bhargav2021,He2022, Rowold2023,Liniger2015}. Due to the non-convexity of most planning problems in racing, different initial guesses can lead to different local optima \cite{Liniger2015}. To find the global optimum, a common approach is to solve multiple OCP in parallel, one for each homotopy class, i.e. overtaking behavior \cite{He2022,Kalaria2021}. The results are then compared in search of the progress-maximizing solution. This approach increases the chances of finding the global optimum at the cost of computational complexity. A different approach that reduces online computational effort is to determine overtaking with a policy learned from offline simulations \cite{Bhargav2021}. While this method is fast, it is not versatile because the calculated policy is only valid for a specific track.

Lim et al. \cite{Lim2018} propose a hierarchical planning approach for traffic scenarios. The behavior is determined with a spatio-temporal graph search, and the solution is used to initialize an \ac{ocp}. This method combines the ability of discrete methods to find solutions close to the global optimum with the precision of trajectories calculated with continuous methods. However, the algorithm is only viable with a low resolution of the graph, resulting in too conservative behaviors for racing.

There are several ways to enforce overtaking behavior in numerical optimization algorithms. Some authors suggest initializing the \ac{ocp} with a trajectory estimate, following the behavior \cite{Buyval2017,He2022,Kalaria2021}. 
Other methods convert the non-convex \ac{ocp} into a convex subproblem by limiting motion to a maneuver envelope so that the behavior of the planned trajectory is more predictable \cite{Ziegler2014,Bender2015,Bhargav2021}. 
This is especially important in racing, where following the optimal behavior is critical.

\subsection{Contributions}

We introduce a hierarchical planning method that extends the local racing line algorithm in \cite{Rowold2023} for multi-vehicle scenarios. Inspired by \cite{Lim2018}, we combine discrete high-level behavioral planning with low-level numerical optimization. This reduces computational complexity compared to \cite{He2022} and \cite{Kalaria2021} and improves flexibility compared to~\cite{Bhargav2021}. The main contributions to the hierarchical approach are as follows:
\begin{itemize}
	\item We propose a behavioral planning step based on spatio-temporal graphs. In contrast to \cite{Lim2018}, temporal planning precedes spatial planning. Progress variants, derived from the previous planning iteration, determine the geometry of spatial planning problems that are solved with low-resolution visibility graphs. 
	\item We adapt the constraints and cost function of the time-\acl{ocp} in \cite{Rowold2023} to generate a feasible trajectory for the generated high-level behavior.
	\item We perform a monte carlo simulation to compare our approach with parallel optimization-based methods and naive overtaking strategies. We analyze the results regarding computation time and driving performance.
\end{itemize}

\section{Methodology}
\begin{figure}
\vspace{0.22cm}
\begin{picture}(100,100)
\put(0,0){\includegraphics[width=\linewidth]{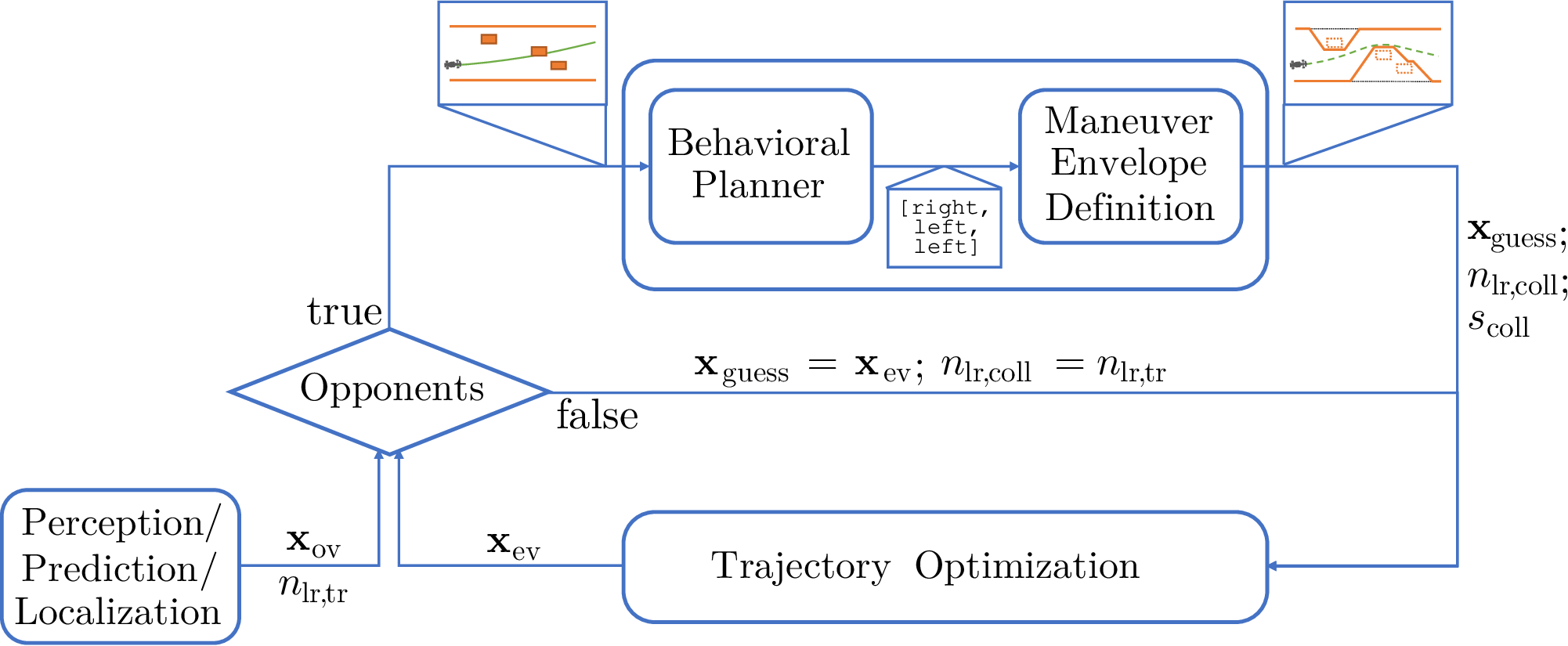}}
\put(180,11){{\ssmall\cite{Rowold2023}}}
\end{picture}
\caption{Overview of the hierarchical planning approach.}
\label{fig:overview}
\end{figure}

Our approach operates in two modes shown in Figure~\ref{fig:overview}:\\(1) Without any opponents in the planning horizon, the trajectory is generated according to \cite{Rowold2023}. A time-\acl{ocp} with a point mass model, constrained by gg-diagrams, is solved for the upcoming track section. In Sections \ref{sec:track_model} and \ref{sec:veh_model}, we will briefly summarize the used track and vehicle model.\\(2) When opponents are present, the first step is to make a behavioral decision, whether to pass opponents on the left or right and to define a corresponding maneuver envelope. These processes are explained in Sections \ref{sec:behavioral_planning} and \ref{sec:maneuver_envelope_definition}. The second step, described in Section \ref{sec:ocp}, is to solve the time-optimal \ac{ocp} with constraints adapted to comply with the determined maneuver envelope.

\subsection{Track Model}
\label{sec:track_model}
\begin{figure}
\centering
\includegraphics[width=0.9\linewidth]{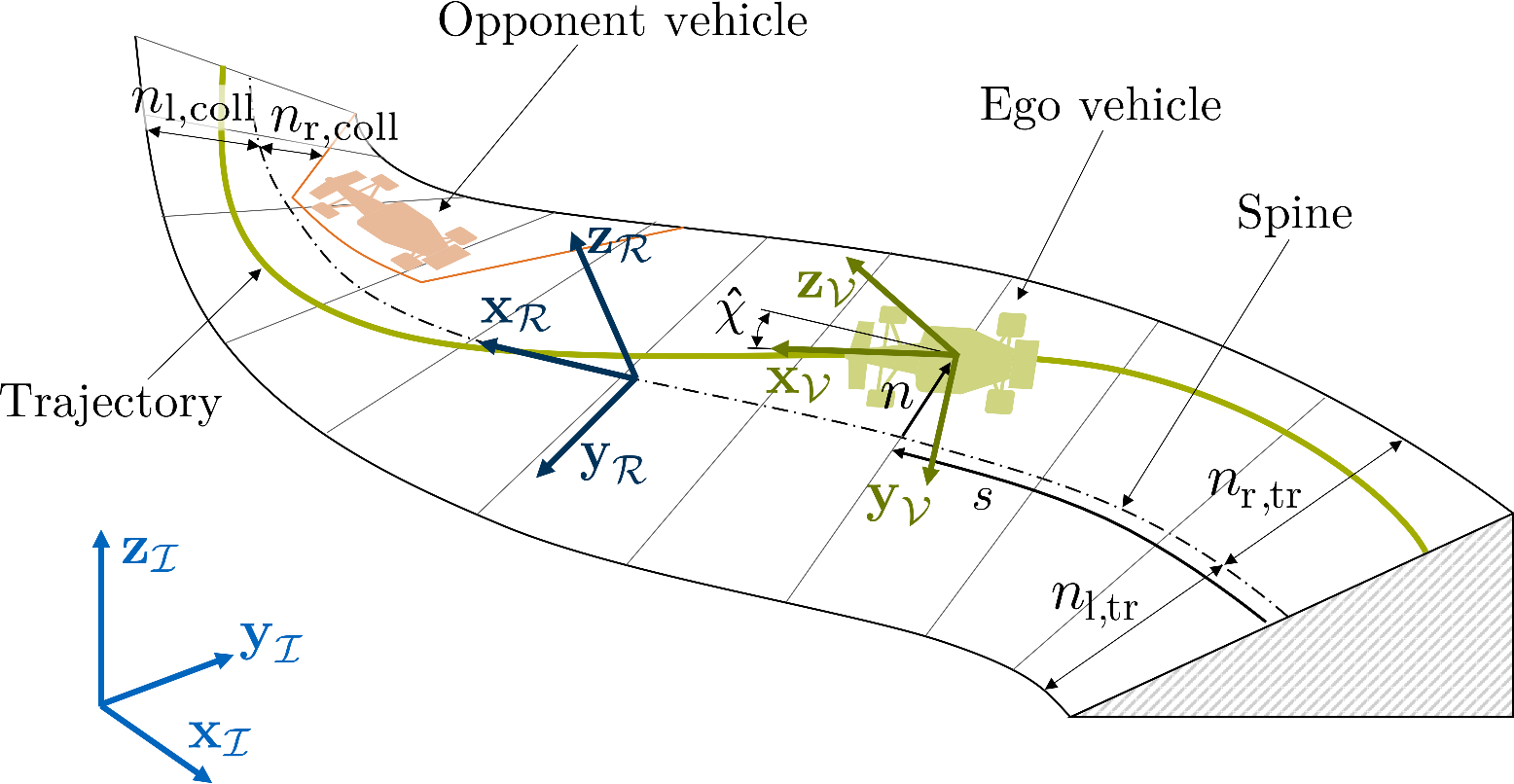}
\caption{\ac{3D} track with road frame $\mathcal{R}$ and velocity frame $\mathcal{V}$.}
\label{fig:track}
\end{figure}

We use the curve-ribbon approach for modeling \ac{3D} tracks, presented in \cite{Perantoni2015}. The road frame $\mathcal{R}$ moves along a \ac{3D} reference curve, called the spine. It defines the road surface, as shown in Figure \ref{fig:track}. The arc length along the spine is denoted as $s$, while $n$ is the lateral displacement in the direction of the y-axis of $\mathcal{R}$. The rotation rate of $\mathcal{R}$ with respect to arc length $s$ is expressed as angular velocity $_\mathcal{R}\boldsymbol{\Omega}_\mathcal{R}= \begin{bmatrix} \Omega_x & \Omega_y & \Omega_z\end{bmatrix}^\top$ in the $\mathcal{R}$-frame. 
For a detailed description of the \ac{3D} track representation, we refer to \cite{Perantoni2015}.

\subsection{Vehicle Model}
\label{sec:veh_model}
Following \cite{Rowold2023}, we use a low-dimensional point mass model to describe the dynamics of the vehicle. The state $\mathbf{x}$ is defined as
\begin{equation}
\mathbf{x}=\begin{bmatrix}
V & n & \hat{\chi} & \hat{a}_\mathrm{x} & \hat{a}_\mathrm{y}
\end{bmatrix}^\top\text{,}
\end{equation}
where $V$ is the velocity and $\hat{\chi}$ is the orientation of the velocity-alinged frame $\mathcal{V}$ relative to the road frame $\mathcal{R}$. The longitudinal and lateral accelerations of $\mathcal{V}$ are given by $\hat{a}_\mathrm{x}$ and $\hat{a}_\mathrm{y}$, respectively. The accelerations are constrained by gg-diagrams according to \cite{Rowold2023}. The longitudinal and lateral jerks $\hat{j}_\mathrm{x}$ and $\hat{j}_\mathrm{y}$ form the input vector
\begin{equation}
\mathbf{u}=\begin{bmatrix}
	\hat{j}_\mathrm{x} & \hat{j}_\mathrm{y}
\end{bmatrix}^\top\text{.}
\end{equation}
With the vertical velocity $w$ and the angular velocity of $\mathcal{V}$ with respect to time $_\mathcal{V}\boldsymbol{\omega}_\mathcal{V}= \begin{bmatrix} \hat{\omega}_x & \hat{\omega}_y & \hat{\omega}_z\end{bmatrix}^\top$, the dynamics are described by
\begin{equation}
	\label{eq:dynamics}
	\dot{\mathbf{x}} = \frac{d\mathbf{x}}{dt}=\mathbf{f}(\mathbf{x}, \mathbf{u}) = \begin{bmatrix}
		\hat{a}_\mathrm{x} - w \hat{\omega}_\mathrm{y}\\
		V\sin(\hat{\chi}) \\
		\frac{\hat{a}_\mathrm{y} + w\hat{\omega}_\mathrm{x}}{V}-\Omega_\mathrm{z}\dot{s}\\
		\hat{j}_\mathrm{x}\\
		\hat{j}_\mathrm{y}
	\end{bmatrix}\text{.}
\end{equation}
\subsection{Behavioral Planning}
\label{sec:behavioral_planning}
\begin{figure}
\vspace{0.22cm}
\centering
\includegraphics[width=0.8\linewidth]{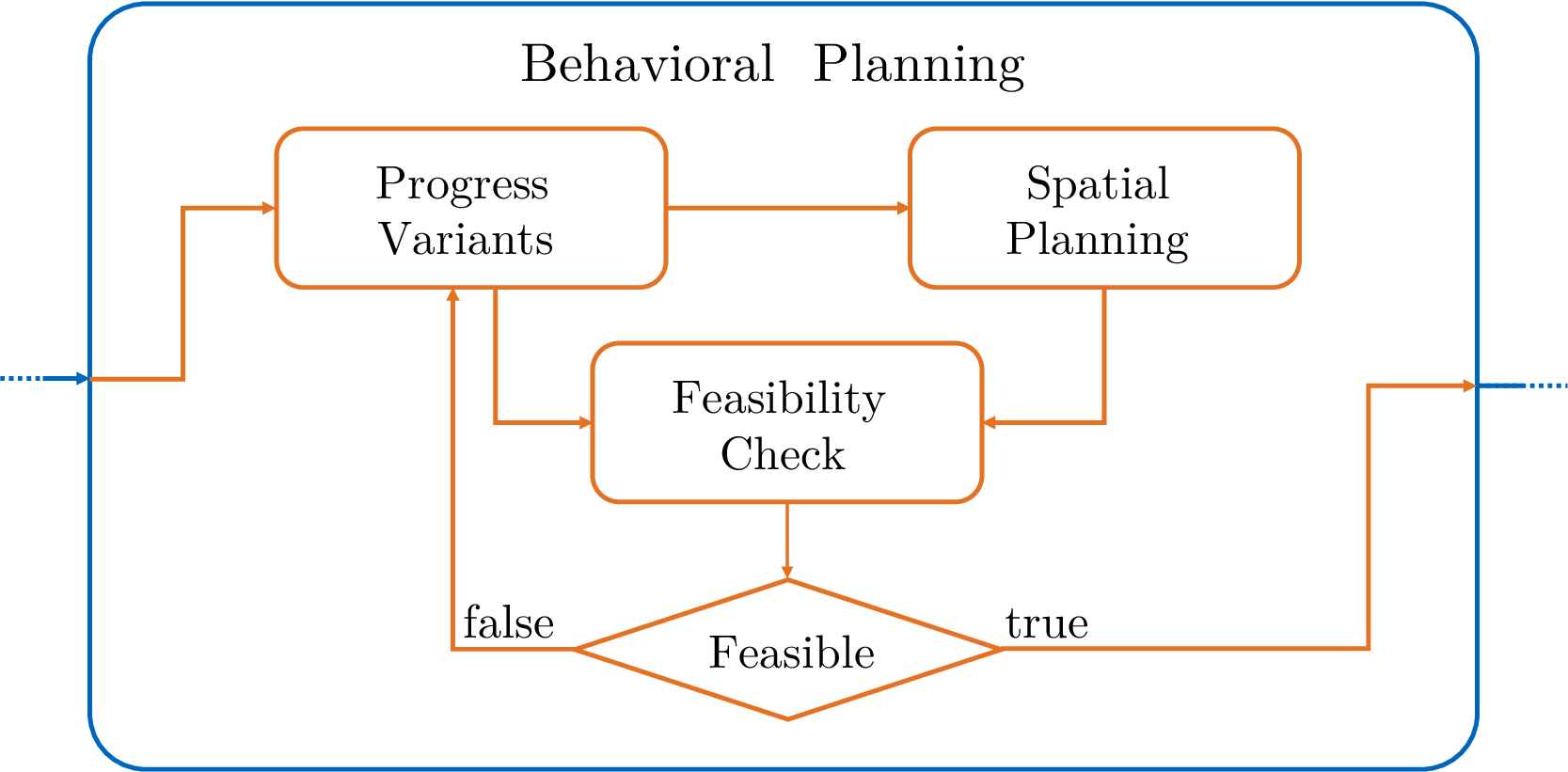}
\caption{Overview of the behavioral planning algorithm.}\label{fig:beh_planning}
\end{figure}
\begin{figure}
\includegraphics[width=\linewidth]{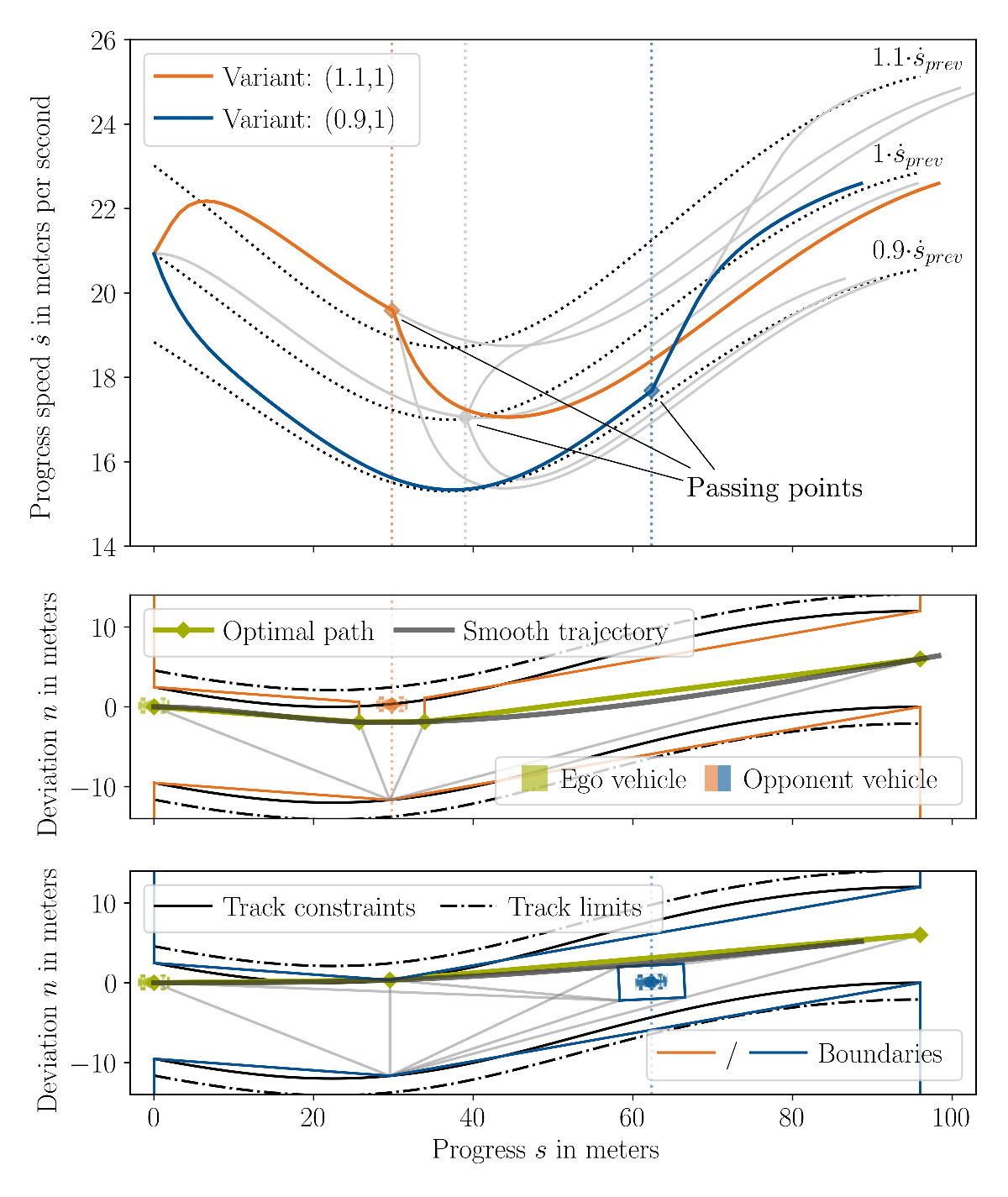}
\caption{Behavioral planning for an example maneuver with a single opponent. The top figure shows the generation of progress variants, while the middle and bottom figures depict the visibility graphs for the variants (1.1,1) and (0.9,1).}\label{fig:example_beh_planning}
\end{figure}
Given the predicted motion of the opponent vehicles, the behavioral planning step approximates an optimal overtaking trajectory and selects the fastest sequence of left or right passing decisions.
Since the times and positions of overtakes depend on the progress of the ego vehicle, this is a spatio-temporal problem. We solve this problem by first sampling progress variants, second finding the optimal path for a given variant, and third checking the feasibility of the resulting trajectory. Following this order allows for the creation of low-resolution visibility graphs that take advantage of the problem geometry. 
\subsubsection{Progress Variants}
Progress variants are generated by following different set speed profiles $\dot{s}_\mathrm{set}$, based on the optimal trajectory of the previous planning iteration $\mathbf{x}_\mathrm{prev}$. To do this, we modulate acceleration with a feedback controller $\ddot{s}=K(\dot{s}_\mathrm{set}-\dot{s})$. Following a certain speed profile $\dot{s}_\mathrm{set}$ determines the $s$-coordinate where the next opponent vehicle is passed. 
At these passing points, the speed profiles branch out by switching to different set speed profiles. The passing points for a single opponent and set speed profiles $\dot{s}_\mathrm{set}\in\{0.9\cdot\dot{s}_\mathrm{prev},$ $1\cdot\dot{s}_\mathrm{prev},$ $1.1\cdot\dot{s}_\mathrm{prev}\}$ are shown in the top diagram of Figure \ref{fig:example_beh_planning}. In the same diagram, we highlight two exemplary progress variants $(1, 1.1)$ and $(0.9, 1)$. 
The first variant follows $1.1\dot{s}_\mathrm{prev}$ at the start of the maneuver and switches to $1\dot{s}_\mathrm{prev}$ after the overtake, while the second one goes from $0.9\dot{s}_\mathrm{prev}$ to $1\dot{s}_\mathrm{prev}$.
For multiple opponents, the aforementioned procedure can quickly result in a large number of progress variants. With three speed profiles and $N$ opponents, $3^{N+1}$ variants are possible. Performing spatial planning, as described in Section \ref{sec:spatial_planning}, for all variants would be too computationally complex. Therefore, we generate the progress variants as needed, beginning with the fastest variant. If the spatial planning step can generate a feasible trajectory for the current variant, the trajectory and corresponding behavior are applied to the numerical optimization. Otherwise, we continue with the next fastest variant. This iterative procedure is visualized in Figure~\ref{fig:beh_planning} and promotes finding the global time-optimal solution. More details on the feasibility checks are given in Section~\ref{sec:feasibility_check}.

\subsubsection{Spatial Planning} 
\label{sec:spatial_planning}
With the passing points of the considered progress variant, a spatial graph can be generated. We utilize visibility graphs. 
These are undirected graphs, connecting all vertices of obstacles with straight edges that do not cross an obstacle \cite{LozanoPerez1979}. Originally, they were developed to find the shortest collision-free path. 
Compared to the spatio-temporal lattice with fixed nodes \cite{Lim2018}, the discretization, based on the corner points of moving obstacle polygons, allows for the generation of short and direct path candidates with a small number of nodes. Considering vehicle dimensions and safety distances, we virtually expand the track boundaries and opponent polygons to avoid collisions when the center point of the ego vehicle is within bounds.
The visibility graphs for the progress variants $(1, 1.1)$ and $(0.9, 1)$ are shown in the two bottom diagrams of Figure~\ref{fig:example_beh_planning}.

An A* search determines the optimal path to minimize the  total travel distance $\sum_i d_i$ and angle deviation $\sum_i|\hat{\chi}_i|$ relative to the spine: $\min\limits_{i}{\sum_i (w_dd_i+w_\chi|\hat{\chi}_i|)}$. The search is guided by a heuristic function $h(P)$, based on the length $d_{\overline{PD}}$ and angle deviation $|\hat{\chi}_{\overline{PD}}|$ of a virtual edge $\overline{PD}$ connecting the current point $P$ to the destination $D$: $h(P)=w_dd_{\overline{PD}}+w_\chi|\hat{\chi}_{\overline{PD}}|$. By discouraging long and weaving paths, the goal is to predict which path is most likely to be feasible for the given progress variant.
We increase the speed of the search algorithm by applying the following simplifications to reduce the number of nodes in the graph: (1) With the help of the Ramer–Douglas–Peucker algorithm \cite{Ramer1972}, we reduce the number of boundary points to a subset of points that approximates the shape. (2) We remove the boundary nodes at the start and end of the planning horizon, as the vehicle would have to drive perpendicular to the spine or in reverse track direction to reach them.
\subsubsection{Feasibility Check}
\label{sec:feasibility_check}
Spatial planning with visibility graphs results in non-continuous curvature profiles, so the unprocessed paths are not feasible. To confirm the suitability of a path and its corresponding overtaking behavior, a cubic spline $f_\mathrm{spline}(s)$ is placed through the path, as seen in Figure~\ref{fig:example_beh_planning}. The smoother path candidate $n_\mathrm{cand}=f_\mathrm{spline}(s_\mathrm{cand}(t))$ is then combined with the considered progress variant $\dot{s}_\mathrm{cand}(t)$ to form the trajectory candidate
\begin{equation}
\label{eq:candidate_trajectory}
\mathbf{x}_\mathrm{cand}=\begin{bmatrix}
V_\mathrm{cand} \\ n_\mathrm{cand} \\ \hat{\chi}_\mathrm{cand} \\ \hat{a}_\mathrm{x,cand} \\ \hat{a}_\mathrm{y,cand}
\end{bmatrix}=
\begin{bmatrix}
\frac{\dot{s}_\mathrm{cand}(1-n_\mathrm{cand}\Omega_\mathrm{z})}{\cos\hat{\chi}}\\
n_\mathrm{cand}\\
\arctan{f_\mathrm{spline}'(s_\mathrm{cand})}\\
\dot{V}\\
V(\omega_z+\dot{\hat{\chi}})
\end{bmatrix}\text{.}
\end{equation}
If the accelerations from the trajectory $\mathbf{x}_\mathrm{cand}$ lie within the gg-diagrams, behavioral planning finishes with the trajectory estimate $\mathbf{x}_\mathrm{guess}=\mathbf{x}_\mathrm{cand}$ and its corresponding behavior. Otherwise, the next slower progress variant is examined.

\subsection{Maneuver Envelope Definition}
\label{sec:maneuver_envelope_definition}
\begin{figure}
\vspace{0.22cm}
\centering
\includegraphics[width=\linewidth]{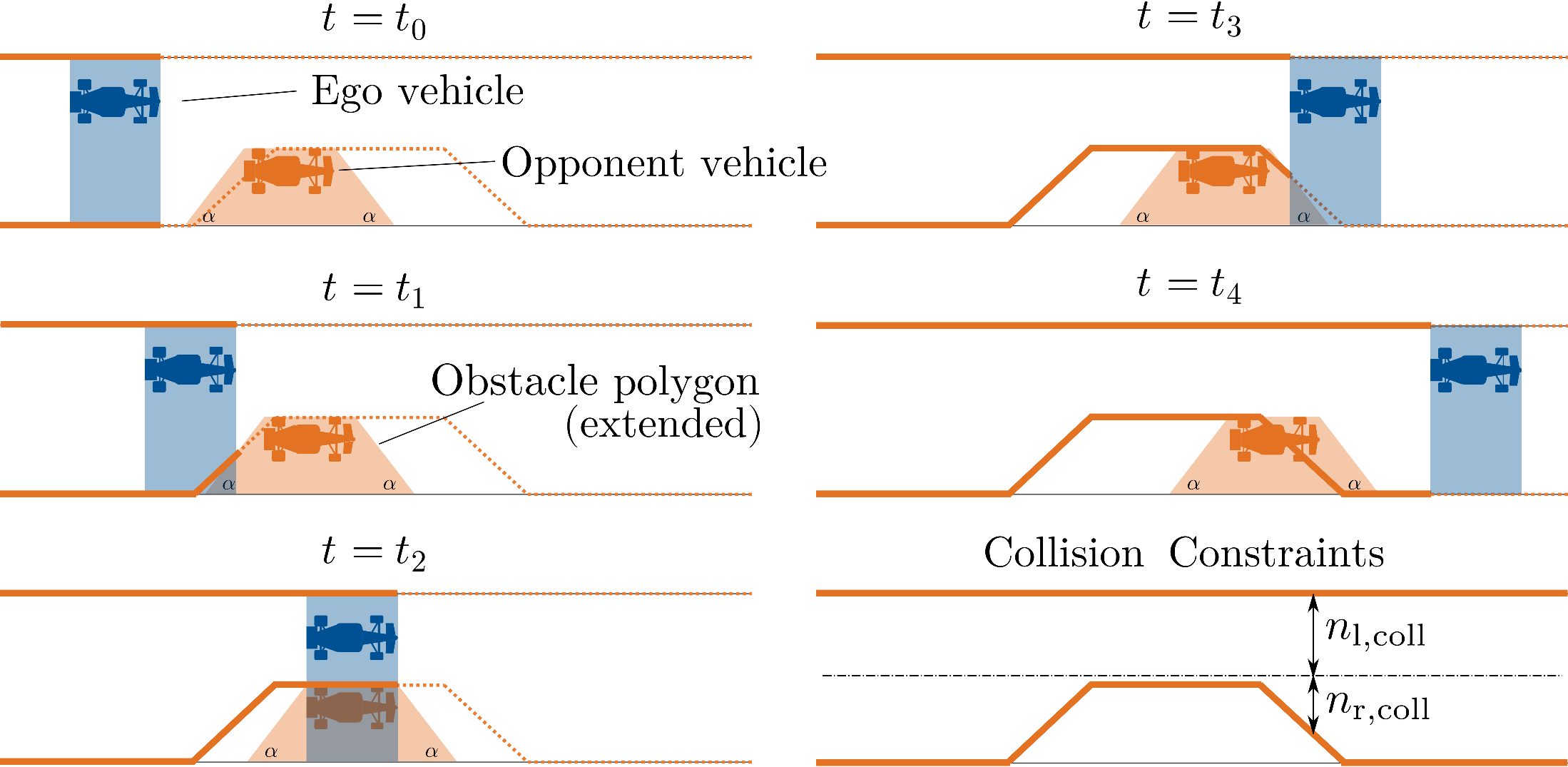}
\caption{Generation of maneuver envelopes.}
\label{fig:coll_constr}
\end{figure}

The maneuver envelope should force the solution of the \ac{ocp} with initialization $\mathbf{x}_\mathrm{guess}=\mathbf{x}_\mathrm{cand}$ to remain in the previously determined optimal behavior class. 

The maneuver envelopes are formed by extending obstacle polygons of the opponents to cover the side of the track where overtaking is suboptimal according to the behavioral planning step. As the vehicle travels along the planning horizon, the resulting spatio-temporal obstacle constraints form a narrowed driving corridor, as depicted in Figure~\ref{fig:coll_constr}. We reduce complexity by combining all obstacle constraints into collision constraints that describe this narrowed driving space.  This is achieved by sampling the lateral restriction $n
\in[n_\mathrm{r,coll}(s),n_\mathrm{l,coll}(s)]$ for the initial guess $\mathbf{x}_\mathrm{guess}$.

While the complexity of the \ac{ocp} is significantly reduced, information gets lost when spatio-temporal constraints are reduced to spatial constraints. As the constraints now only depend on the distance, they can influence the vehicle speed solely through the feasible curvatures in the narrowed driving space. 
To make the vehicle slow down when the gap for overtaking is too small, we re-introduce the spatio-temporal component by adding a constraint on vehicle progress $s<s_\mathrm{coll}+V_\mathrm{coll}t$. This longitudinal constraint acts like a wall moving at the average speed of the obstacle.

\subsection{Optimal Control Problem}
\label{sec:ocp}

Following \cite{Rowold2023}, the second step of our hierarchical approach solves an \ac{ocp} parametrized by $s$ for a constant spatial planning horizon $s\in[s_0,s_e]$. 
By using numerical optimization, we can calculate fast trajectories that are not limited by discretization.
The cost function \eqref{cost} consists of three terms: a time optimality term, a term that smooths the acceleration profile by minimizing jerk, and a slack term that ensures that the soft constraints on velocity and vehicle position are fulfilled. The \ac{ocp} is defined as
\begin{subequations}
\begin{align}
\begin{split}\label{cost}
\min_{\mathbf{x},\textbf{u}} \quad & \int_{s_0}^{s_e} \frac{1}{\dot{s}} + 
\textbf{u}^\top\textbf{R}\textbf{u}+
\boldsymbol{\epsilon}^\top\textbf{S}\boldsymbol{\epsilon}\: ds
\end{split}\\
\begin{split}
\text{s.t.}\quad & \mathbf{x}'=\textbf{f}(\mathbf{x},\textbf{u})\frac{1}{\dot{s}}\label{eom}
\end{split}\\
\begin{split}
V-\epsilon_V\leq V_{max}\:\text{with}\:\epsilon_V\geq 0\label{v_constr}
\end{split}\\
\begin{split}
\text{(9a), (9b), (9c) in \cite{Rowold2023}}\label{acc_constr1}
\end{split}\\
\begin{split}
n_\mathrm{r,tr}+d_\mathrm{s}\leq n \leq n_\mathrm{l,tr}-d_\mathrm{s}\label{n_track}
\end{split}\\
\begin{split}
-\frac{\pi}{2}\leq\hat{\chi}\leq\frac{\pi}{2}\label{chi_constr}
\end{split}\\
\begin{split}
	n-\epsilon_{n_{\mathrm{l,coll}}}+d_\mathrm{s}\leq n_\mathrm{l,coll}\:\text{with}\:\epsilon_{n_{\mathrm{l,coll}}}\geq 0\label{n_coll1}
\end{split}\\
\begin{split}
	n+\epsilon_{n_{\mathrm{r,coll}}}-d_\mathrm{s}\geq n_\mathrm{r,coll}\:\text{with}\:\epsilon_{n_{\mathrm{r,coll}}}\geq 0\label{n_coll2}
\end{split}\\
\begin{split}
	s-\epsilon_{s\text{,coll}}\leq s_\mathrm{coll}\:\text{with}\:\epsilon_{s\text{,coll}}\geq 0\label{s_coll}
\end{split}
\end{align}
\end{subequations}
with
\begin{equation}
\begin{aligned}
\textbf{R}= & \begin{bmatrix}
w_{j,\text{x}} & 0\\
0 & w_{j,\text{y}}
\end{bmatrix},\\
\boldsymbol{\epsilon}= & \begin{bmatrix}1 & \epsilon_V& \epsilon_{n_{\mathrm{l,coll}}}& \epsilon_{n_{\mathrm{r,coll}}} \epsilon_{s_\mathrm{coll}}\end{bmatrix}^\top,\\
\textbf{S}=&\left[\begin{array}{@{\mkern0mu} c @{\mkern1mu} c @{\mkern1mu} c @{\mkern1mu} c @{\mkern1mu} c @{\mkern3mu}}
0&\frac{w_{\epsilon,V,1}}{2}&\frac{w_{\epsilon,n_{\mathrm{l,coll}},1}}{2}&\frac{w_{\epsilon,n_{\mathrm{r,coll}},1}}{2}&\frac{w_{\epsilon,s_\mathrm{coll},1}}{2}\\
\frac{w_{\epsilon,V,1}}{2}&w_{\epsilon,V,2}&0&0&0\\
\frac{w_{\epsilon,n_{\mathrm{l,coll}},1}}{2}&0&w_{\epsilon,n_{\mathrm{l,coll}},2}&0&0\\
\frac{w_{\epsilon,n_{\mathrm{r,coll}},1}}{2}&0&0&w_{\epsilon,n_{\mathrm{r,coll}},2}&0\\
\frac{w_{\epsilon,s_\mathrm{coll},1}}{2}&0&0&0&w_{\epsilon,s_\text{coll},2}
\end{array}\right]\text{.}
\end{aligned}\nonumber
\end{equation}
Constraint \eqref{eom} enforces the equations of motion in \eqref{eq:dynamics}. Using the diamond interpolation method presented in \cite{Rowold2023}, we limit the combined accelerations in \eqref{acc_constr1}. The vehicle is kept a safety margin $d_s$ away from the track boundaries in \eqref{n_track}. Inequality \eqref{chi_constr} prevents driving in reversed track direction. The aforementioned restrictions are hard constraints that have to be satisfied for a solution to exist.

However, there are cases where such a strict definition of constraints might be disadvantageous regarding the robustness of the solver. E.g., the speed limit is not safety-critical and can be violated for short periods of time. The soft velocity constraint is realized by the slack variable $\epsilon_V$ in \eqref{v_constr}. Similarly, the maneuver envelopes from Section \ref{sec:maneuver_envelope_definition} are realized as soft constraints with \eqref{n_coll1}--\eqref{n_coll2}. With hard constraints, the result, if feasible, would be too conservative because the uncertainty of the opponent's prediction increases with distance. Following \cite{Kerrigan2000}, our slack variables have linear and quadratic terms in the cost function. These are realized by the matrix $\mathbf{S}$. If the initialization of the \ac{ocp} violates one of the soft constraints $x_0 \nleq x_\mathrm{max}$, the corresponding slack variable $\epsilon_x$ is initialized with the value of the excess $\epsilon_{x,0}=x_0-x_\mathrm{max}$. Within and between the planning steps, the violation is gradually decreased and eventually eliminated. The linear and quadratic weights $w_{x,1}$, $w_{x,2}$ determine how hard the violations are penalized and are therefore used for adjusting the softness of the constraints. This is especially useful for the collision constraints \eqref{n_coll1}--\eqref{s_coll}. Here, the slack weights $w_{\epsilon,i_\mathrm{coll},j}(s)$ for $i\in\{n_\mathrm{l},n_\mathrm{r},s\}$ and $j\in\{1,2\}$ are defined  as a function of progress with parameters $w_{\epsilon,i_\mathrm{coll},j,s0}$ and $w_{\epsilon,i_\mathrm{coll},j,se}$ ($w_{\epsilon,i_\mathrm{coll},j,s0}>w_{\epsilon,i_\mathrm{coll},j,s\text{e}}$)
\begin{equation}
w_{\epsilon,i_\mathrm{coll},j}(s)=w_{\epsilon,i_\mathrm{coll},j,s\text{e}}\left(\frac{w_{\epsilon,i_\mathrm{coll},j,s0}}{w_{\epsilon,i_\mathrm{coll},j,s\text{e}}}\right)^\frac{s_e-s}{s_e-s_0}\text{.}
\end{equation}
Large slack weights close to the ego vehicle ($s\approx s_0$) reduce the likelihood of collisions. Meanwhile, low weights at the end of the planning horizon make the solver more stable in the presence of large and sudden changes in the predicted vehicle position. For the velocity constraint \eqref{v_constr}, we use constant slack weights $w_{\epsilon,V,1}$ and $w_{\epsilon,V,2}$.

As long as the behavior, determined by the high-level planning step in Section~\ref{sec:behavioral_planning}, remains the same, we initialize the \ac{ocp} with the solution of the previous optimization $\mathbf{x}_\mathrm{guess}=\mathbf{x}_\mathrm{prev}$. 
If the behavior changes, the smoothed trajectory, passing the feasibility check in Section~\ref{sec:feasibility_check}, is used as a new initial guess for the \ac{ocp} $\mathbf{x}_\mathrm{guess}=\mathbf{x}_\mathrm{cand}$.

\section{Results}
To validate and evaluate the hierarchical planning approach, we perform randomized simulations with three vehicles on the Modena race track. All simulations are calculated on an Intel Core i7-5600U CPU. The planning horizon is set to  $H=s_\mathrm{e}-s_0=\SI{300}{\meter}$ and progress variants are generated with the set speed factors $\{0.5,1,2\}$. The slack weights 
$w_{\epsilon,n_\mathrm{coll},1,s0/se}=50/25$, $w_{\epsilon,n_\mathrm{coll},2,s0/se}=5/2.5$, $w_{\epsilon,s_\mathrm{coll},1,s0/se}=20/10$ and $w_{\epsilon,s_\mathrm{coll},2,s0/se}=2/1$
are selected for collision-free overtaking. The definition of all other parameters was guided by \cite{Rowold2023}. At the beginning of each simulation, two opponent vehicles are positioned at random within a range of \SI{120}{\meter} in front of the ego vehicle. Once the ego vehicle has overtaken both opponents and opened a gap of \SI{50}{\meter}, the simulation is stopped. Each initial configuration is tested with the following four approaches:
\begin{enumerate}
	\item The hierarchical approach presented in this paper
	\item A pseudo-parallel optimization approach
	\item Overtaking only on the left of the opponents
	\item Overtaking only on the right of the opponents
\end{enumerate}
The pseudo-parallel approach is based on the parallel optimization from Section \ref{sec:related_work}. For each behavior class, we specify a maneuver envelope and solve an \ac{ocp}. To initialize the different \ac{ocp}s, we generate path candidates with cubic splines.  
These paths start at the current vehicle position, pass through points of the obstacle polygons and finish on the track spine at the end of the planning horizon.
We initialize the \ac{ocp} with $\mathbf{x}_\mathrm{guess}=\mathbf{x}_\mathrm{prev}$ for the behavior class equal to the previous solution of the planning algorithm. For a comparison with our single-core hierarchical approach, we solve all \ac{ocp}s sequentially. We name this single-core variant of parallel optimization pseudo-parallel optimization.

\begin{figure}
\includegraphics[width=\linewidth]{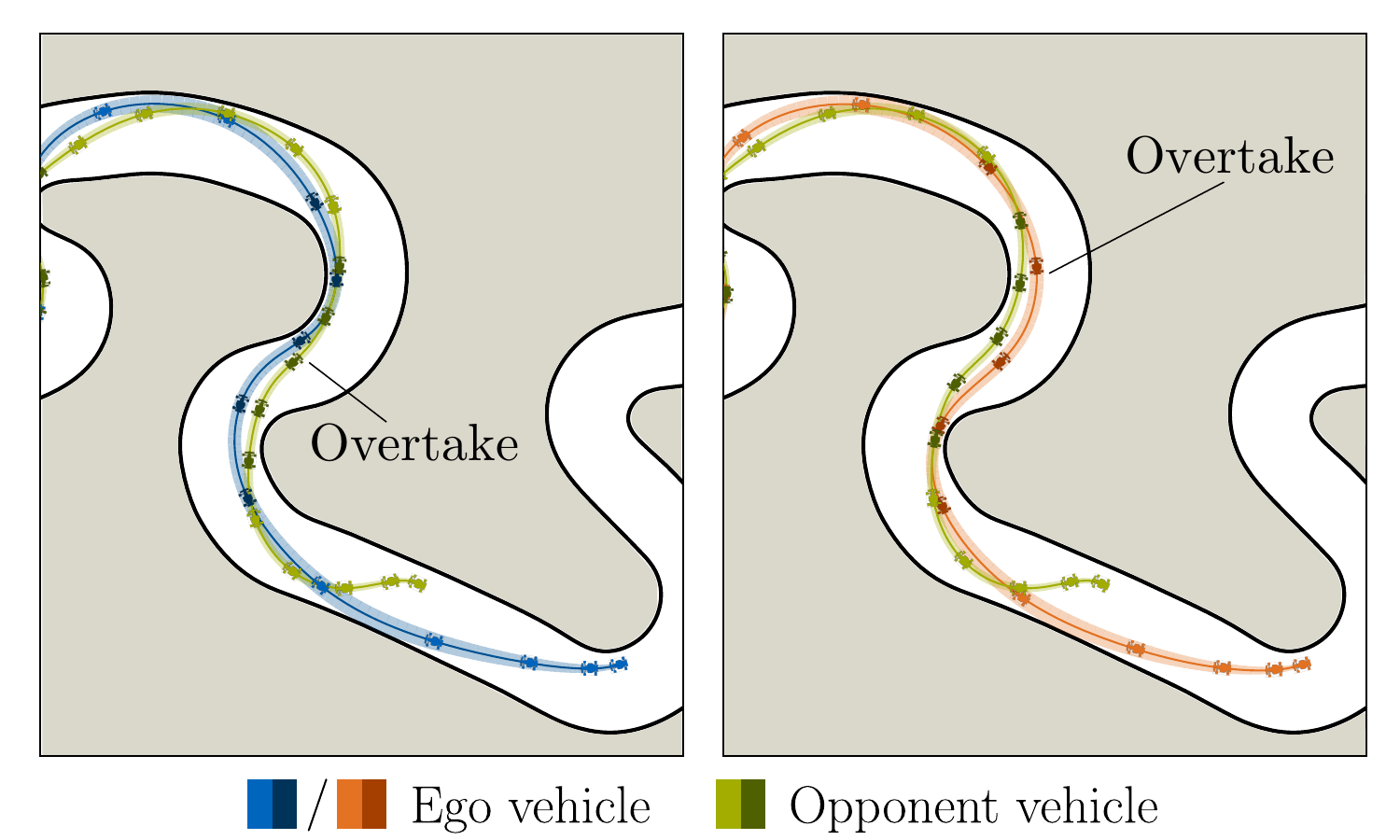}
\caption{Example for left overtaking behavior (left) and right overtaking behavior (right). The lines show the paths of the vehicles and line thickness indicates speed. The vehicle positions are plotted at intervals of $\approx \SI{1.2}\second$.}\label{fig:ex_overtake}
\end{figure}

Exemplary overtakes for approaches 3) and 4) are shown in Figure~\ref{fig:ex_overtake}. When overtaking on the left is specified, the vehicle passes on the outside of the first corner. For the overtake on the right, the vehicle accelerates less at the beginning to overtake when a gap opens up on the outside of the second turn. In this scenario, the left behavior class results in an earlier overtake compared to the right one.

\begin{figure}
\includegraphics[width=\linewidth]{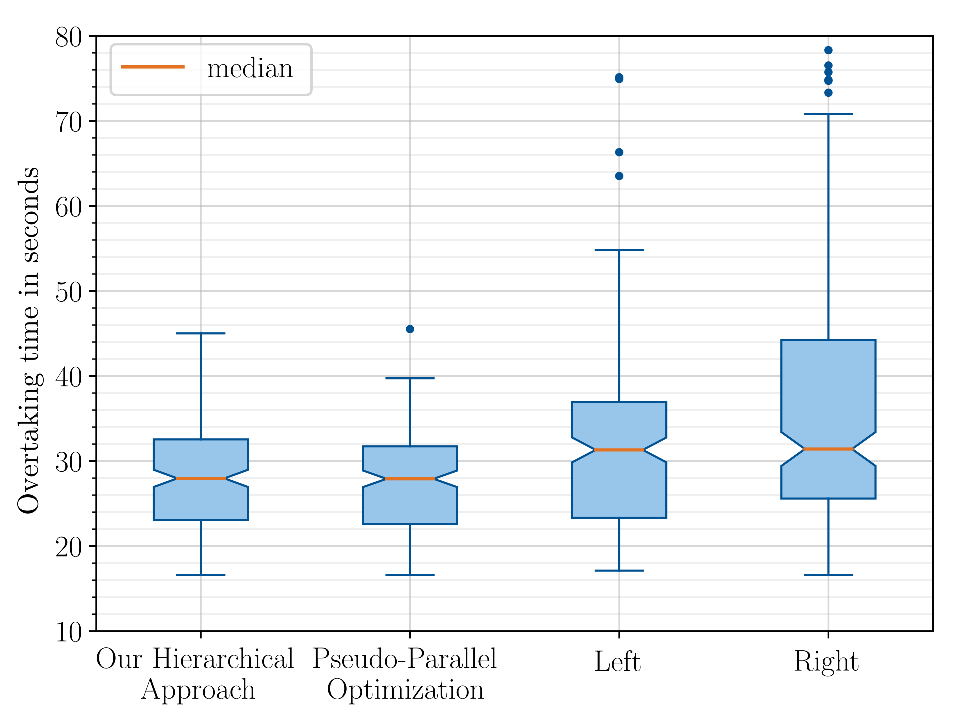}
\caption{Time required for overtaking two opponent vehicles based on 216 randomized overtaking simulations. 
The median overtaking performance of our hierarchical approach is similar to the pseudo parallel optimization. Compared to that, the fixed strategies generally result in slower overtakes.}\label{fig:overtaking_time}
\end{figure}

Figure \ref{fig:overtaking_time} shows the duration of the overtaking simulations for approaches 1)--4). Compared to the first two methods, the fixed behaviors produce significantly longer and more inconsistent overtaking times. This shows that there are multiple local optima and that it is beneficial to select the correct behavior before solving the \ac{ocp}. The overtaking times from our hierarchical approach are similar to the pseudo-parallel approach that assesses all behavior classes in detail. Thus, we deduce that our proposed method selects the optimal behavior in the majority of cases.

\begin{figure}
\includegraphics[width=\linewidth]{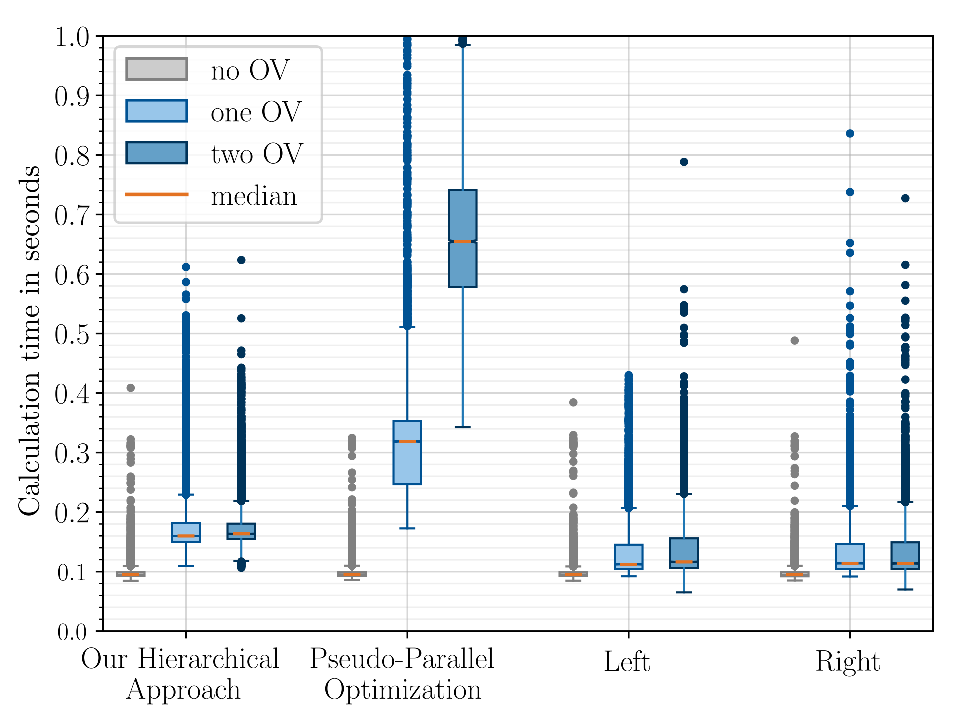}
\caption{Calculation time for different maneuvers and numbers of opponent vehicles (OV) based on 216 simulations with approximately 300 steps per simulation. 
%
The hierarchical approach shows slightly higher computation times than methods without behavioral planning. However, in contrast to the pseudo-parallel approach, the computation time does not increase exponentially with the number of opponents. Thus, it is more scalable.
%
}\label{fig:calc_time}
\end{figure}

The ego vehicle considers the vehicles within its planning horizon. To evaluate the scalability of the planning approaches 1)--4), we assess the calculation times for different numbers of opponents within the planning horizon. The results are shown in Figure~\ref{fig:calc_time}. For the constant overtaking behaviors 3) and 4), the calculation times are similar and largely unaffected by the addition of opponents.
For our hierarchical approach, there is a \SI{0.6}{\second} jump when introducing the first opponent. Adding a second opponent does not result in any further increase. Thus, computational complexity appears to scale well with the number of opponents. 

Conversely, calculation time increases exponentially for the pseudo-parallel approach. Instead of planning in two steps, it optimizes a trajectory for every possible behavior combination. As there are $2^N$ combinations for $N$ opponents (e.g. $N=2 \rightarrow [\text{ll,lr,rl,rr}]$), the calculation time doubles for every added opponent. Going from zero to one opponent, the computation time increases even more, as only one of the \ac{ocp}s is initialized with the previous solution when there are opponents. The other initial guesses are farther from their corresponding optima, and the \ac{ocp}s take longer to converge.

\section{Conclusion}
In this paper, we introduced a hierarchical planning approach for racing with multiple opponent vehicles. We demonstrated that our approach with low-resolution visibility graphs for spatio-temporal planning is computationally efficient and has overtaking performance similar to parallel optimization. Therefore, the method is to be preferred when the computational resources are limited to a single core.

\bibliographystyle{IEEEtran}
\bibliography{IEEEabrv,bibtex/literature}

\end{document}